\documentclass[11pt]{article} 

\usepackage{graphicx}   
\usepackage{amsmath, amsfonts}  
\usepackage{url}        
\usepackage{bm}         
\usepackage[ruled,vlined]{algorithm2e}  
\usepackage{multirow}   
\usepackage{xfrac}      
\usepackage{caption}
\usepackage{subcaption}
\usepackage{setspace}
\usepackage{fix-cm}
\usepackage[left]{lineno} 
\usepackage[margin=3cm]{geometry}
\title{Swarm Oracle:\\ Trustless Blockchain Agreements through Robot Swarms}

\author{
    Alexandre Pacheco$^{1,4,^{\ast}}$
    Hanqing Zhao$^{2}$,
    Volker Strobel$^{1}$,
    Tarik Roukny$^{6}$,
    Gregory Dudek$^{3}$,\\
    Andreagiovanni Reina$^{1,4,5}$,
    Marco Dorigo$^{1}$
}

\date{} 

\begin{document}

\maketitle

{\small \noindent
$^{1}$IRIDIA, Universit\'{e} Libre de Bruxelles, Brussels, Belgium\\
$^{2}$Department of Electrical and Computer Engineering, Universit\'{e} Laval, Québec, Canada\\
$^{3}$School of Computer Science, McGill University, Montr\'{e}al, Canada\\
$^{4}$Centre for the Advanced Study of Collective Behaviour, Universit\"{a}t Konstanz, Germany\\
$^{5}$Department of Collective Behaviour, Max Planck Institute of Animal Behavior, Konstanz, Germany\\
$^{6}$Faculty of Economics and Business, KU Leuven, Belgium\\
$^{\ast}$To whom correspondence should be addressed; \texttt{alexandre.melo.pacheco@gmail.com} \\
}

\begin{abstract}

Blockchain consensus, rooted in the principle ``don't trust, verify'', limits access to real-world data, which may be ambiguous or inaccessible to some participants.
Oracles address this limitation by supplying data to blockchains, but existing solutions may reduce autonomy, transparency, or reintroduce the need for trust.
We propose Swarm Oracle: a decentralized network of autonomous robots---that is, a robot swarm---that use onboard sensors and peer-to-peer communication to collectively verify real-world data and provide it to smart contracts on public blockchains.
Swarm Oracle leverages the built-in decentralization, fault tolerance and mobility of robot swarms, which can flexibly adapt to meet information requests on-demand, even in remote locations.
Unlike typical cooperative robot swarms, Swarm Oracle integrates robots from multiple stakeholders, protecting the system from single-party biases but also introducing potential adversarial behavior. 
To ensure the secure, trustless and global consensus required by blockchains, we employ a Byzantine fault-tolerant protocol that enables robots from different stakeholders to operate together, reaching social agreements of higher quality than the estimates of individual robots.
Through extensive experiments using both real and simulated robots, we showcase how consensus on uncertain environmental information can be achieved, despite several types of attacks orchestrated by large proportions of the robots, and how a reputation system based on blockchain tokens lets Swarm Oracle autonomously recover from faults and attacks, a requirement for long-term operation.

\end{abstract}
\section{Introduction}

Blockchain technology facilitates the creation and governance of public digital resources through peer-to-peer collaboration over the internet. Bitcoin~\cite{Nak2008:techreport} exemplifies this as the first successful public monetary ledger, enabling censorship-resistant transactions worldwide. The key ideas behind Bitcoin's success were later adopted and generalized by Ethereum~\cite{But2014:techreport}, a blockchain envisioned as a ``world computer'' that hosts smart contracts---programs written in a Turing-complete language and executed collectively by the Ethereum network.
To operate and scale globally, blockchain systems are designed to be decentralized, transparent, and trustless, relying on large networks of nodes to independently verify that new blocks follow consensus rules and that user transactions are valid. However, this design imposes a fundamental limitation: blockchains cannot securely incorporate information unless it is algorithmically verifiable by every node.
This limitation extends to smart contracts which rely on accurate real-world data to support decision-making, such as exchange rates for decentralized finance, seismic activity measurements for disaster insurance, or environmental indicators for managing smart cities and natural resources~\cite{antonopoulos2018mastering}. 

External entities that provide data to blockchains are called \emph{oracles}, a term borrowed from Alan Turing’s ``oracle-machines''---hypothetical devices that can solve problems beyond the limits of algorithmic computation~\cite{turing1939systems}. However, relying on oracles to obtain accurate real-world data reintroduces elements of trust and centralization, potentially undermining blockchain's core properties such as automation, censorship resistance, and transparency. This issue is known as blockchains' ``oracle problem''~\cite{Cal2020:information}.
As an illustrative example, consider a smart contract that distributes rewards to local landowners for maintaining a sustainable forest (e.g., using funds originating from the sale of carbon credits). In principle, such a contract could implement transparent, immutable rules that guarantee landowners direct and equitable access to funds, independent of shifting politics, opaque bureaucracies and financial intermediaries.
However, to determine rewards, the contract requires reliable information about the state of the forest: Are biodiversity levels increasing? At what rate are trees being felled? Has a fire occurred? 
Prospective blockchain projects could, for instance, obtain this data from a trusted environmental agency or autonomous software that processes satellite images---both of which acting as oracles.

To reduce the need for trust, decentralized oracle networks such as Chainlink~\cite{2021chainlink} aggregate data from multiple software oracles that retrieve and process online information---such as the satellite images in our example, but also asset prices and news events. 
Beniiche~\cite{beniiche2020study} proposes a taxonomy of existing oracle solutions based on their underlying trust models, ranging from authoritative to decentralized consensus-based approaches, and on the nature of their participants (e.g., software, hardware, and even human).
In contrast with software oracles, which rely on online sources, hardware oracles can collect data directly from the physical environment using sensors. A prominent example is Helium~\cite{Haleem2024Helium}, whose IoT network provides wireless coverage while gathering geolocation and connectivity data for on-chain applications. 
However, despite growing demand for transparent and trustworthy data, existing decentralized oracles, hardware ones in particular, lack the autonomy, flexibility, and scalability required for most applications. 

In this paper, we propose Swarm Oracle, a decentralized oracle network composed of autonomous robots that act as hardware oracles, directly sensing and processing data from the physical world. By leveraging swarm robotics' peer-to-peer interactions, decentralization, and self-organization~\cite{Dorigo:2014}, Swarm Oracle can potentially scale or adapt robots' tasks to flexibly meet the demands of real-world applications. 
Even when composed of simple robots equipped with inexpensive, noisy or biased sensors, Swarm Oracle harnesses the wisdom of the crowd~\cite{Galton1907VoxPopuli,ValHamDor2015:aaai-video} through a consensus protocol, aggregating data from multiple robots to reach agreements that are more accurate and reliable than individual robots' sensor readings. 
Swarm Oracle can, for example, monitor fish stocks, wildfires, traffic conditions, or trash deposits, providing important information required by smart contracts to enact transparent regulations to manage shared resources~\cite{hassan2021decentralized}.
In this context, Swarm Oracle functions as a Decentralized Physical Infrastructure Network (DePIN)~\cite{lin2024decentralized}, retrieving and serving real-world data to blockchains on demand, as part of the growing decentralized economy.

We envision Swarm Oracle as a collective of robots from different stakeholders, a prerequisite for a \textit{trustless} oracle network that prevents biases during the design, manufacturing, or programming of the robots. 
Achieving consensus in a mobile multi-robot system with unreliable and sparse communication is a major challenge, further complicated by the possibility that stakeholders hold conflicting interests or that malicious actors attempt to manipulate on-chain data for personal gain. 
Even under these adversarial conditions, Swarm Oracle consensus must be both \textit{fault-tolerant} and \textit{global}---otherwise, discrepancies among the robots could compromise the integrity and reproducibility of the blockchain’s deterministic state.

Unfortunately, these requirements are not met by typical swarm robotics consensus methods, where robots gradually adjust their individual beliefs (e.g., through local averaging of sensor readings~\cite{ValFerDor2017:frontiers}) to converge toward an \textit{approximate} collective value, possibly applying local rejection of extreme values to enhance robustness (e.g., the W-MSR rule \cite{LeBZhaKouSun2013:comm}). While some strategies can withstand adversarial robots~\cite{leblanc2011adversaries, sundaram2012adversaries, leblanc2012lowcomp}, security hinges on robots' remaining sufficiently connected to honest robots. In sparsely connected robot swarms, these approaches may fail when networks partition---especially if malicious robots target specific neighborhoods. Most critically, they allow for persistent (small) discrepancies in opinions across the swarm, falling short of Swarm Oracle’s requirements.

\begin{figure}
    \centering
    \includegraphics[width=\linewidth]{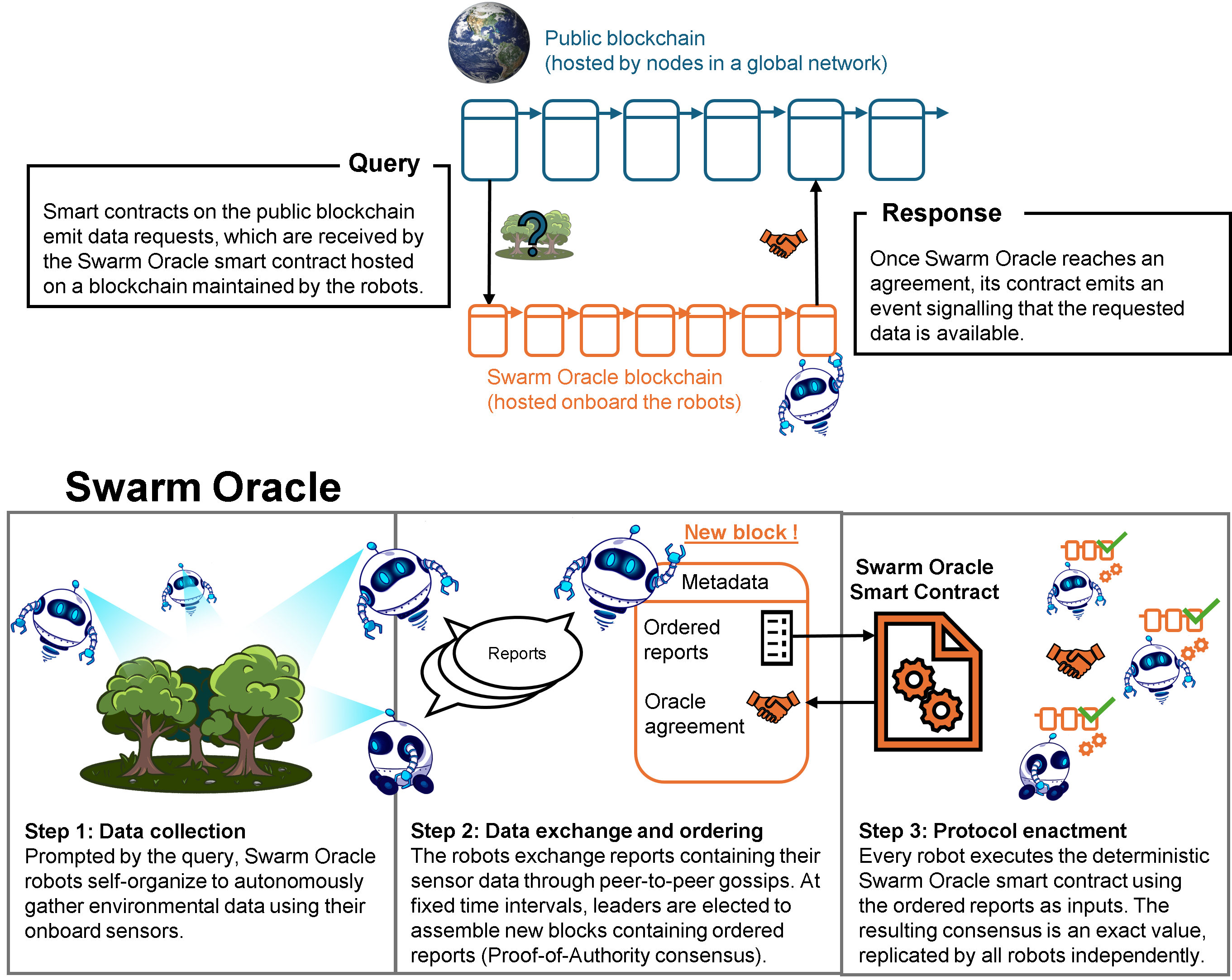}
    \caption{Deployment of the Swarm Oracle as a service that delivers reliable data to a public blockchain. In this deployment scenario, the protocol is implemented as a smart contract on a dedicated second-layer blockchain maintained by the robots. 
    By hosting the contract on a dedicated blockchain tailored to their capabilities, the robots can reach a deterministic global (i.e., swarm-wide) agreement from the aggregation of individual robots' data, with lower latency and transaction costs than a public blockchain deployment.}
    \label{fig:demonstration}
\end{figure}

To reach trustless, fault-tolerant and global consensus without requiring external infrastructure, we implement Swarm Oracle through a blockchain hosted by the robots themselves. This second-layer blockchain (also called ``sidechain'') is adapted to the limitations of the robots, employing a lightweight consensus protocol to avoid the high energy demands of protocols based on proof-of-work~\cite{Nak2008:techreport}. The Swarm Oracle protocol, which includes data aggregation methods, is deployed as a smart contract on this blockchain, as illustrated in Figure~\ref{fig:demonstration}.
The Swarm Oracle smart contract interacts with external infrastructure through events, receiving queries from public blockchains and publishing consensus results once agreements are reached. A general taxonomy on information exchange methods between oracle networks and smart contracts is available in~\cite{edu_oracles}.
While our framework centers on blockchain as both the application and the deployment vehicle, the trustworthy information produced by Swarm Oracle can also support off-chain decision processes, particularly when transparency and impartiality are required, such as applications for non-profit or government institutions. The Swarm Oracle protocol could equally be deployed as a smart contract on a public blockchain, though this could introduce transaction costs, especially when swarm sizes are large or when robots collect high volumes of data~\cite{DorPacReiStr2024naturereviewstech}.

Previous research~\cite{StrCasDor2018:aamas,StrCasDor2020:frontiers, PacStrDor2020:ants, StrPacDor2023:sciencerobotics} showed how robot swarms can benefit from blockchains to detect inconsistent or extreme values reported by some robots, and how to use blockchain tokens as participation credentials to protect the system from ``double-spending'' and ``Sybil'' attacks.\footnote{Attacks where participants send conflicting information to different peers ( ``double-spending''), and where they generate false identities (``Sybil'').}
However, the ``oracle agreement'' in these studies, in which robots vote and agree on the fraction of white floor tiles, offered no formal guarantees against attacks, and only a few malicious behaviors were studied. A small group of coordinated robots could potentially synchronize their votes to exploit the system, thereby gaining cryptotokens and voting power. 
Zhao et al.~\cite{ZhaPacStr-etal2023:iros} presented a blockchain-based Byzantine fault-tolerant algorithm, 
however, this preliminary work lacked validation on physical robots, leaving open questions about the practicality of the approach and its ability to withstand coordinated attacks under realistic constraints, without incurring excessive delays or communication overhead.

In this work, we deploy a Byzantine fault-tolerant consensus 
protocol on a swarm of 12 physical robots, to form an experimental Swarm Oracle. We stress-test the system at its theoretical limit of fault tolerance: when one third of the robots coordinate attacks to compromise the Swarm Oracle's \textit{safety} and \textit{liveness} properties~\cite{lamport1977proving}.\footnote{The safety property states that \textit{bad things won't happen}---in our case, that the robots will not agree on incorrect information. The liveness property states that \textit{good things will happen}---in our case, that agreements are reached within feasible time and resource constraints.}
We also integrate a reputation system that enables Swarm Oracle to self-heal from faults and attacks, as this is a crucial requirement for continuous and autonomous operations. Unlike previous reputation systems~\cite{PacStrDor2020:ants, StrPacDor2023:sciencerobotics, ZhaPacStr-etal2023:iros} that penalized only robots providing unreliable information, Swarm Oracle also safeguards liveness by penalizing robots that become inactive, either due to accumulating faults or attacks.
In our experimental scenario, robots use onboard cameras to identify the colors of landmarks in the environment. This scenario reflects real-world conditions where robots have varying levels of accuracy and individual biases in their sensing, as color perception is highly influenced by environmental and hardware variability (e.g., camera calibration and lighting conditions). We perform an extensive experimental study using real robots (approximately 22 hours of runtime) and simulated robots (approximately 278 hours of runtime), showcasing how the Swarm Oracle provides correct agreements and recovers from attacks and faults.

\section{Results}
\label{sec:app}

\subsection{Experimental Scenario}

\begin{figure}
\centering
\includegraphics[width=0.5\linewidth]{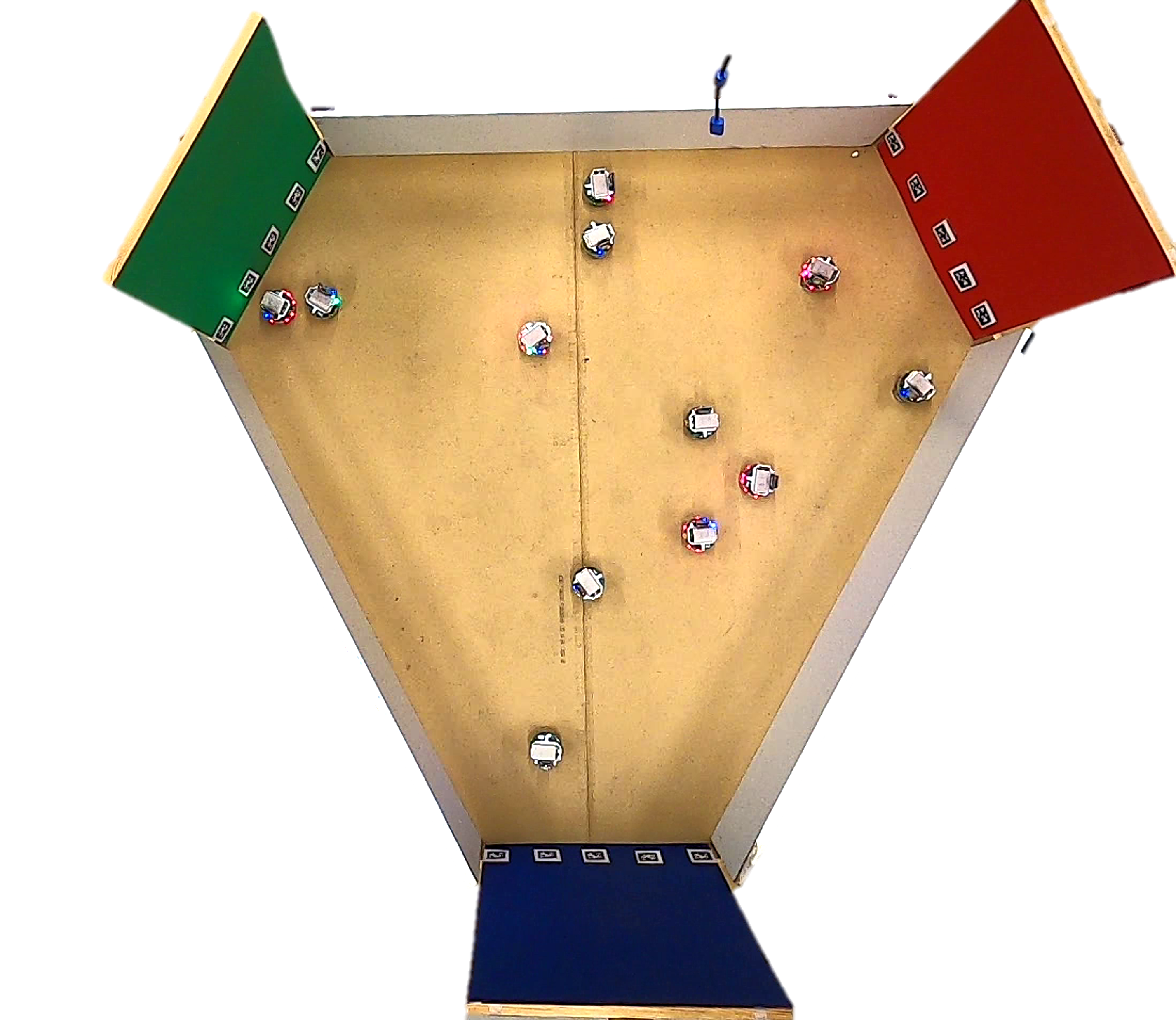}
\caption{The arena contains three colored landmarks that can be detected by the robot cameras from a distance (potentially from any position in the arena if there is a clear line of sight) and used for navigation. Each landmark has AprilTags, readable from a short distance (approximately 10~cm), which encode ``not valuable'' on green and blue landmarks, and ``valuable'' on red.}
\label{fig:environment}
\end{figure}

Our experimental scenario, shown in Figure~\ref{fig:environment}, contains three colored landmarks (red, green, and blue), each marked with AprilTags indicating whether they are ``valuable'' or not. In this study, the red landmark is designated as valuable, while green and blue are not. The oracle query under consideration is: ``What are the RGB color values of the valuable landmark?'' 
This setup illustrates a potential real-world deployment of a Swarm Oracle, composed of a swarm of 12 mobile robots that explore the environment and monitor its physical properties using onboard sensors and peer-to-peer communications.
Each robot is equipped with an inexpensive monocular camera used to detect the colors of the landmarks and to read the AprilTags. However, color readings are highly noisy due to occlusions, sensor quality, and light conditions. Additionally, individual differences in each robot’s color perception introduce systematic biases. As a Swarm Oracle, robots combine their readings to compute an accurate consensus value for the RGB color of the landmark (a three-dimensional array of continuous values). Robots can navigate toward visible color blobs, but can only determine whether a landmark is valuable or not by reading the AprilTags at close range (approximately 10 cm). Adversary robots can use the green and blue landmarks as distractors to deliver some of their attacks.

\subsection{The Swarm Oracle}
The design and properties of Swarm Oracle are presented in detail in Methods. Here, we provide a brief, intuitive overview in the context of our experimental scenario.

Robots move toward large color blobs they detect and, within 10~cm of an AprilTag, they stop to sample RGB values. These values (or \emph{observations}) are then reported to nearby robots (i.e., robots not farther away than approximately 15~cm). Each report additionally contains the sending robot’s identity, a deposit of reputation tokens, and a vote indicating the value of the resource.
A blockchain, maintained by the robots in their peer-to-peer network, hosts Swarm Oracle’s consensus protocol as a smart contract, orders reports, and tracks the reputation tokens held by each robot---which add up to a total circulating supply of $T$ tokens.

The smart contract aggregates reports into clusters via k-means clustering, and computes proposals which are the centroids of the clusters. A new cluster is created when the distance between an observation and existing proposals exceeds a \emph{clustering threshold}~$R$.
 
To send reports, robots must deposit a fixed fraction of the reputation tokens it owns, which we call \emph{deposit quota} $K \in \mathopen]0,1\mathclose]$. Swarm Oracle's consensus protocol requires that proposals accumulate a pool of deposited tokens larger than $\frac{2}{3}KT$ before a consensus decision is made, therefore, the value $K$ regulates the maximum number of proposals that Swarm Oracle can run in parallel.
Once $\lfloor K^{-1} \rfloor$ proposals are awaiting a decision, no more clusters are generated and reports that do not match existing ones are dropped.

Once the $\frac{2}{3}KT$ pool of tokens is reached, the decision is made in favor of the majority votes, individually weighted by deposits. If the threshold exceeds $\frac{2}{3}$, an adversary with less than $\frac{1}{3}$ of the tokens could cause a deadlock simply by not sending any reports. If the threshold is below $\frac{2}{3}$, the adversary could force arbitrary agreements with under $\frac{1}{3}$ of the tokens (i.e., half of $\frac{2}{3}$) by acquiring a majority stake in a proposal. As such, requiring a threshold of $\frac{2}{3}$ of $KT$ tokens guarantees both liveness and safety, as long as adversaries control no more than one third of the tokens~\cite{lamport1982bft, Castro1999}.

Autonomous Swarm Oracle deployments require recovery mechanisms to prevent faults accumulating over time. To this end, we implement a \textit{reputation system} which rewards reputation tokens to robots in the majority group and penalizes the others, according to Equation~\ref{eq:reward} in Methods. Each time a consensus is reached, the deposits in reports contradicting the majority are redistributed to the majority (improving safety), and a fixed amount of tokens---established by the issuance constant $I_c$---is awarded to the majority (improving liveness).
By using reputation tokens as both participation credentials and weights for future contributions, Swarm Oracle heals itself from the damage caused by Byzantine faults and attacks.

\subsection{Byzantine faults and attacks}

We use the term \emph{Byzantine} to refer to both faults and attacks, emphasizing that Swarm Oracle makes no distinction between the two in its design.

To evaluate Swarm Oracle's fault tolerance, we systematically introduce \emph{Byzantine attackers} into the swarm. The attackers are robots with deliberately modified behaviors designed to degrade Swarm Oracle's safety and/or liveness. In each experiment, all attackers employ the same modified behavior, since a coordinated strategy (collusion) maximizes their chances of success. We investigate four distinct attack types, each targeting a critical aspect of the system’s operation:

\begin{itemize}
    \item \textbf{Safety attack}: The attackers report green and blue landmarks as ``valuable'' and red as ``not valuable''.
    \item \textbf{Liveness attack}: The attackers abstain from voting. 
    \item \textbf{Combined attack}: The attackers combine the two previous strategies: they report for green and blue landmarks, but abstain from red.
    \item \textbf{Physical attack}: The attackers attempt to disrupt other robots' camera readings by physical blocking their line of sight and activating blue LEDs.
\end{itemize}

Robots also exhibit \emph{Byzantine faults} that occur unintentionally, i.e., without explicit fault injection. Faults can affect safety, for example when obstructions or unusual lighting conditions lead to incorrect proposals. They also affect liveness, for example when robots become stuck due to low-quality obstacle sensors or wheel actuators. Unlike attacks, Byzantine faults are considered transient, since we expect that the robots eventually recover and resume normal operations. If persistent faults occur (e.g., broken motors, damaged sensors or disconnected cables), we halt the experiment and repair the affected robots to prevent faults from accumulating and biasing results over time.

\subsection{Experiments}

The experiments were conducted in two batches: \textit{short-run} and \textit{long-run}. 
In the short-run experiments, performed on a physical swarm of 12 Pi-puck robots~\cite{MilJoyHil-etal2017:iros}, we study the effect that the clustering threshold $R$ and the number of Byzantine attackers $f$ have on the quality and costs of agreements (Table~\ref{tab:sr}). The experiments terminate once the first agreement on the valuable landmark is reached. We analyze the costs, defined as the time and number of reports to reach consensus, and the consensus error, measured as a Euclidean distance in RGB space.

\begin{figure}
\centering
\includegraphics[width=0.66\linewidth]{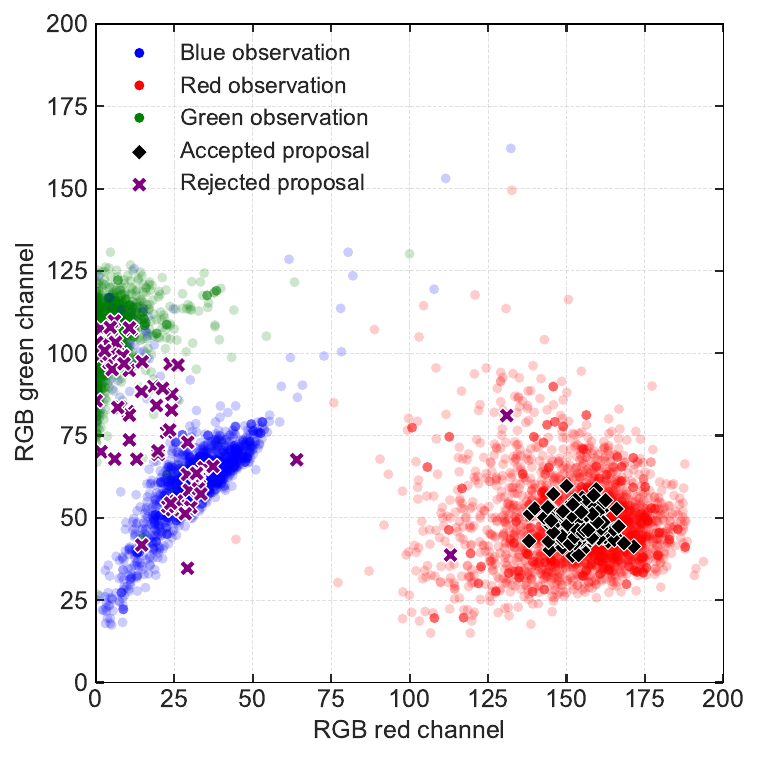}
\caption{Short-run observations and proposals point cloud (two dimensions shown for clarity), with a clustering threshold $R=60$. The proposals initiated due to attacker observations are rejected with 100\% success rate (purple crosses). Some noisy red observations due to unintentional faults led to two rejections. Increasing $R$ would make the system more inclusive of noisy observations, but also increase the variance in accepted proposals.}
\label{fig:rg_cloud}
\end{figure}

In the long-run experiments, simulated in ARGoS~\cite{argos}, we let multiple agreements occur, evaluating how our reputation system performs over time. We vary two key design parameters: the deposit quota $K$ and the issuance constant $I_c$, as shown in Table \ref{tab:lr}.
The simulations accurately replicate key physical parameters of the environment and the robots, including physical collisions, field-of-view occlusions, motion, and communication range. To simulate noisy color perception accurately, each time a robot detects a landmark, it samples at random an RGB observation of that landmark made by a corresponding real robot. A depiction of all RGB observations and Swarm Oracle agreements generated during the short-run experiments is shown in Figure~\ref{fig:rg_cloud}.

\begin{table}
    \centering
    \normalsize

\begin{tabular}{cl|cccc|ccccc}
\multicolumn{1}{l}{} &  & \multicolumn{4}{c|}{\textbf{\begin{tabular}[c]{@{}c@{}}Clustering\\ Threshold $R$\\ (with $\mathbf{f=0}$)\end{tabular}}} & \multicolumn{5}{c}{\textbf{\begin{tabular}[c]{@{}c@{}}\# of Byzantine\\ Attackers $f$\\ (with $\mathbf{R=60}$)\end{tabular}}} \\
\multicolumn{1}{l}{} &  & 20 & 40 & 60 & 80 & 0 & 1 & 2 & 3 & 4 \\ \hline
\multirow{5}{*}{\textbf{\begin{tabular}[c]{@{}c@{}}Type\\ of\\ Attack\end{tabular}}} 
& Safety   & - & - & - & - & - & 10 & 10 & 10 & 10 \\
& Liveness & - & - & - & - & - & 10 & 10 & 10 & 10 \\
& Combined & - & - & - & - & - & 10 & 10 & 10 & 10 \\
& Physical & - & - & - & - & - & 10 & 10 & 10 & 10 \\
& No attack & 10 & 10 & 10* & 10 & 10* & - & - & - & - \\ \hline
\end{tabular}
            
    \caption{Short-run experiments (real robots), showing the number of trials per configuration. The star symbol (*) indicates when parameter configurations are repeated. In total, we performed 200 trials over a total duration of $22\,\mathrm{h}\,32\,\mathrm{m}$.}
    \label{tab:sr}
\end{table}
\begin{table}
    \centering
    \normalsize
    \begin{tabular}{cl|cc|cc}
    \multicolumn{1}{l}{} &  & \multicolumn{2}{c|}{\textbf{\begin{tabular}[c]{@{}c@{}}Deposit \\ Quota K\\  (with $\mathbf{I_c = KT_0}$)\end{tabular}}} & \multicolumn{2}{c}{\textbf{\begin{tabular}[c]{@{}c@{}}Issuance \\ Constant $\mathbf{I_c}$\\  (with $\mathbf{K = \sfrac{1}{3}}$)\end{tabular}}} \\
    \multicolumn{1}{l}{} & \textbf{} & 1 & $\sfrac{1}{3}$ & 0 & $KT_0$ \\ \hline
    \multirow{4}{*}{\textbf{\begin{tabular}[c]{@{}c@{}}Type\\ of\\ Attack\end{tabular}}} & Safety & 40 & 30* & 30 & 30* \\
    & Liveness & 40 & 30* & 30 & 30* \\
    & Combined & 40 & 30* & 30 & 30* \\
    & Physical & 40 & 30* & 30 & 30* \\ \hline
\end{tabular}%
    \caption{Long-run experiments (in simulation), showing the number of trials per configuration. The star symbol (*) indicates when parameter configurations are repeated. In all long-run configurations we use $f=4$ Byzantine attackers and $R=60$ clustering threshold. In total, we performed 400 trials over a total duration of $278\,\mathrm{h}\,52\,\mathrm{m}$.}
    \label{tab:lr}
\end{table}

\subsubsection{Short-run}

\begin{figure}
\centering
\includegraphics[width=0.66\linewidth]{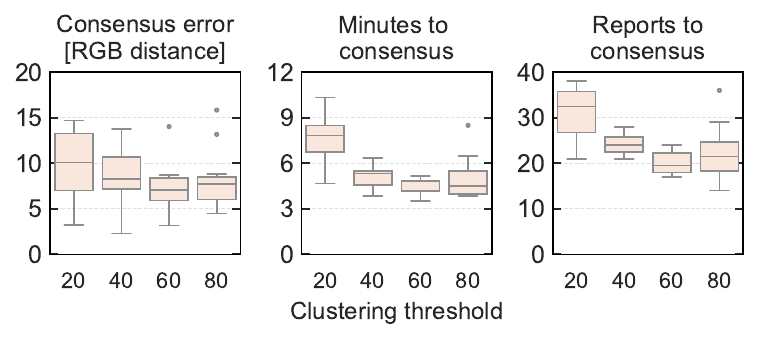}
\caption{Short-run consensus error (given as a Euclidean distance in the RGB space) and costs when varying the clustering threshold. The plots suggest that, in this application, there is an ideal value for which the error and costs are lowest. We use this value ($R=60$) for the other experimental configurations.
}
\label{fig:short_R}
\end{figure}

Figure~\ref{fig:short_R} shows that setting the clustering threshold $R$ too low leads to an increase in both consensus error and costs. This happens because the reports from robots observing the same landmark may be assigned to different proposals, increasing the time and number of reports required to reach the agreement. However, the creation of multiple proposals does not compromise liveness or safety because proposals are eventually accepted. On the other hand, if $R$ is set too high, there is a risk that observations of distinct landmarks will be incorrectly clustered together. This may pose a safety threat, for example, if noisy observations of red are matched to green or blue proposals, resulting in an incorrect consensus. It is therefore safer to adopt a conservative threshold, though not excessively so, in order to maintain reasonable costs. Based on these results, we fix the clustering threshold at $R=60$ for all subsequent experiments.

\begin{figure}
\centering
\includegraphics[width=0.66\linewidth]{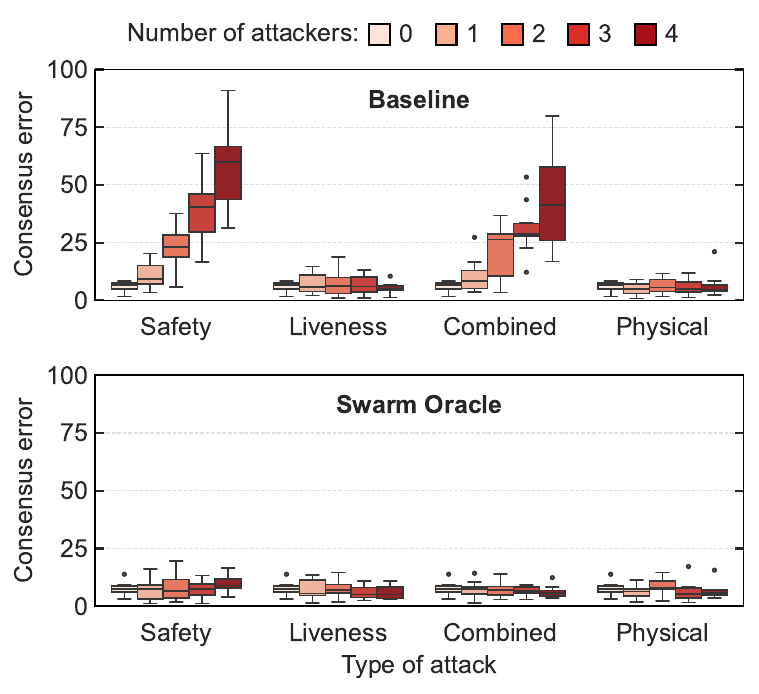}
\caption{Short-run consensus error given as an Euclidean distance in the RGB space.}
\label{fig:error_sr}
\end{figure}

Having established a reasonable clustering threshold, we study how the Swarm Oracle performs when we introduce attacking robots. In Figure~\ref{fig:error_sr}, the baseline (the arithmetic average of all reports) performs poorly, with high error under safety and combined attacks due to the integration of incorrect observations. It also shows high variance, since in some runs attackers detect colors less often and thus send fewer reports. 

Swarm Oracle, with fewer than $4$ attackers (one third), is capable of rejecting all incorrect reports and arriving at an estimate that is as accurate as the attack-free results (the consensus error stays below 20 RGB distance units). Although there is a chance that unintentional faults compound with attacks and cause incorrect agreements, in particular when there are $4$ attackers, this did not occur in the short-run experiments (it did however occur in the long-run experiments). 


\begin{figure}
\centering
\includegraphics[width=0.66\linewidth]{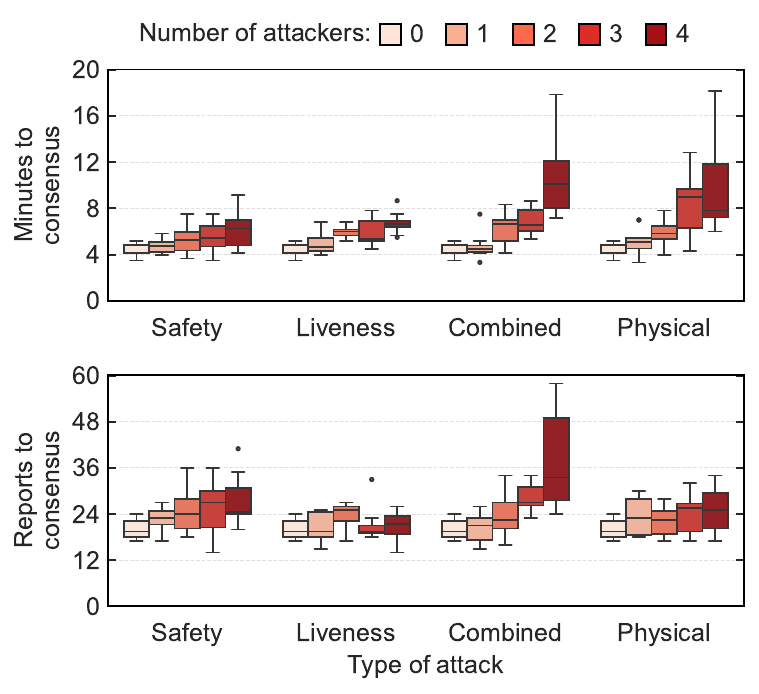}
\caption{Short-run costs to reach consensus in terms of time and number of reports.}
\label{fig:costs_sr}
\end{figure}

Figure~\ref{fig:costs_sr} shows that consensus costs, both in terms of time and reports, increase with the number of attackers. However, the reasons behind these increases are fundamentally different for each type of attack. 

\begin{figure}
\centering
\includegraphics[width=0.66\linewidth]{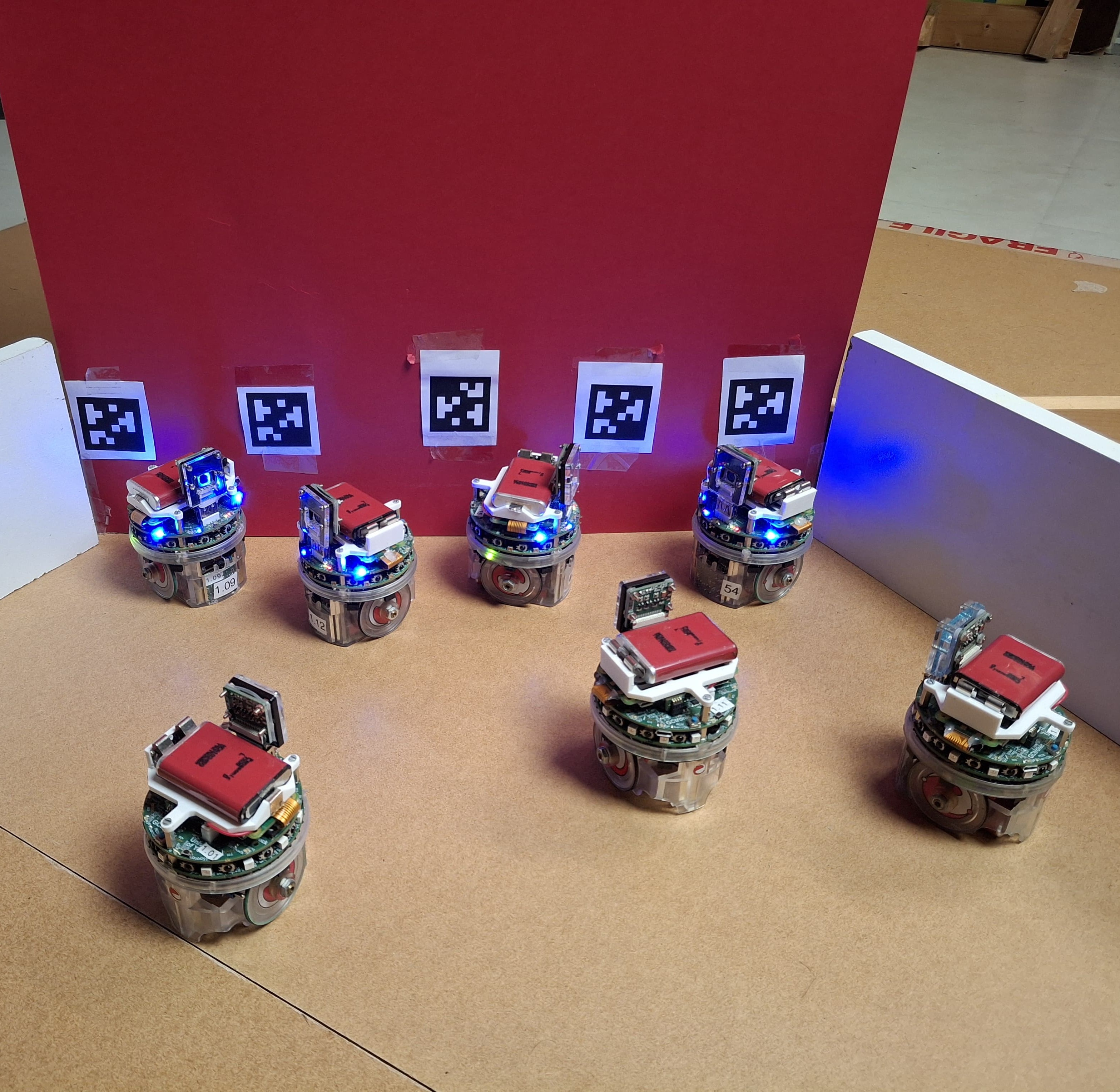}
\caption{Bodywall formed by the physical attackers.}
\label{fig:bodywall}
\end{figure}

Facing \textit{safety attacks}, the non-attacking robots must divide their efforts between validating correct proposals and rejecting incorrect ones. As a result, both the time and number of reports required to reach consensus increase.
In contrast, during \textit{liveness attacks}, attackers do not submit incorrect proposals, so the number of reports remains constant.  However, the time to reach consensus still increases with the number of attackers, at a similar rate to that of safety attacks because, unlike safety attacks---where attackers’ rejection votes help to meet the $\frac{2}{3}$ threshold---liveness attacks make the system more vulnerable to unintentional faults that prevent robots from reporting (e.g., getting stuck or not locating the panel due to obstructions or traffic congestion).
The \textit{combined attack} compounds the effect of both previous attacks, resulting in the highest overall costs in terms of time and reports. For this reason, our reputation system penalizes each attack type independently, leading combined attackers to lose reputation faster (see long-run results).
Finally, in \textit{physical attacks} the attackers attempt to induce faults in other robots, to deteriorate liveness (by physically blocking them) and safety (by partially obstructing the landmark and turning blue LEDs on to degrade perception quality). 
However, once physical attackers find the red landmark, they remain static and do not initiate as many incorrect proposals. This results in lower reporting costs compared to the combined attack. Conversely, the time to consensus matches that of the combined attack, since physical attackers can prevent other robots from reaching the landmark. This is most evident with 3 and 4 attackers since they can effectively form a body-wall in front of the red landmark (Figure~\ref{fig:bodywall}).

\subsubsection{Long-run}

Through a series of longer experiments, we evaluate the effectiveness of our reputation system, described by Equation~\ref{eq:reward}. We aim to study the redistribution dynamics and verify that, by the end of the experiments, the non-faulty robots own most of the reputation tokens and that the system has recovered from faults and attacks, thereby regaining both security margins and operational efficiency.

\begin{figure}
\centering
\includegraphics[width=0.66\linewidth]{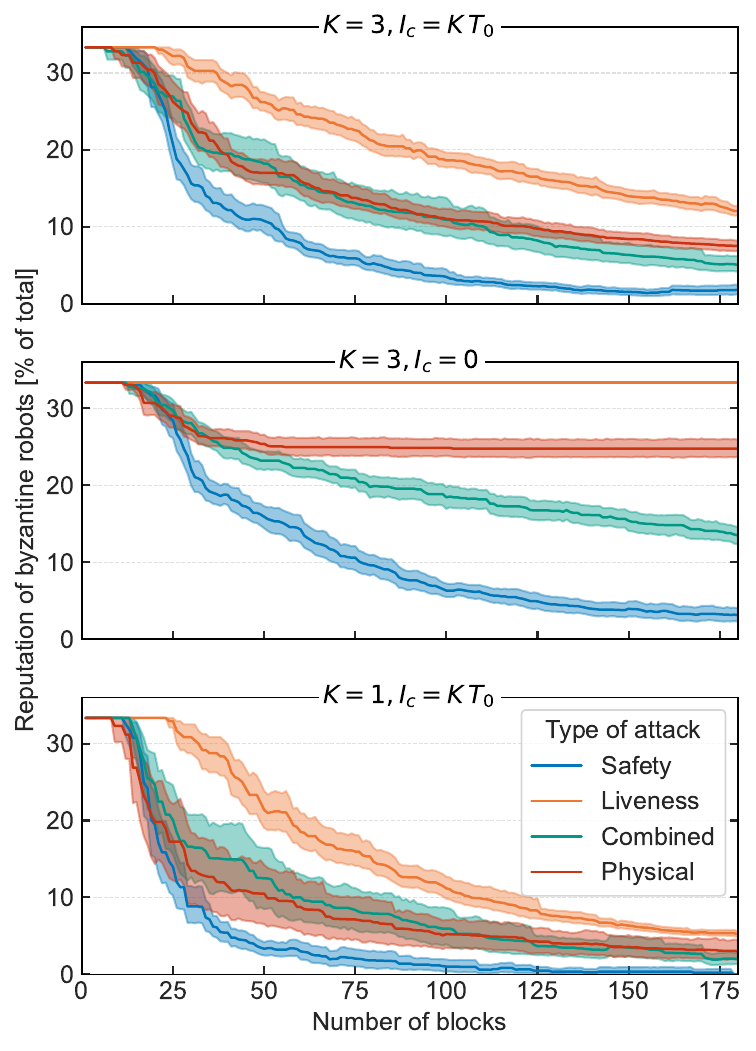}
\caption{Long-term evolution of the fraction of reputation tokens held by attacking robots. Reputation updates occur as new blocks are added to the blockchain, approximately every 10 seconds.}
\label{fig:reputation_lr}
\end{figure}

In the experiment shown in Figure~\ref{fig:reputation_lr}, each of the $12$ robots initially receives an equal share of reputation tokens, therefore the 4 attackers together begin with $33.3\%$ of the total. We chose to evaluate the system under this limit condition, where Swarm Oracle may fail if any additional robot behaves incorrectly, even if unintentionally. This failure occurred in 5 of the 400 long-run trials, in which the attackers ultimately acquired more than $60\%$ of the reputation tokens. These trials are excluded from the plots, as they represent an operating regime beyond Swarm Oracle’s intended design.
Importantly, once an initial correct agreement is reached, reputation tokens are redistributed, making it increasingly difficult for the attackers to succeed.

Although all faulty robots may lose reputation, the attackers are the most penalized, ending with a significantly lower proportion of the total tokens. The rate at which they lose tokens changes with the parameters regulating the reputation system, namely, the deposit quota $K$ and the issuance constant $I_c$. 
With $K=1$, robots can only work on one consensus proposal at a time, leading to faster agreements and redistribution of tokens. Our choice of $K=\frac{1}{3}$ for the majority of the experiments initially leads to slower agreements. However, it allows robots to physically distribute themselves in the environment when contributing to different proposals, resulting in more efficient and robust operation, and in more consistent intervals between subsequent agreements.
The issuance constant $I_c$ establishes how many tokens are created when a new agreement is reached. While safety attackers lose their deposits after a successful vote---according to the second term in Equation~\eqref{eq:reward}---, liveness attackers are only penalized when $I_c > 0$. In this case, the first term in Equation~\eqref{eq:reward} dilutes the stake of non-participating robots by issuing new tokens.
Finally, different forms of attacks lead to different rates of token loss. Since safety attackers do not abstain from voting, agreements occur more frequently, leading to a faster redistribution of tokens. The combined and physical attackers are penalized by both terms in Equation~\eqref{eq:reward}, however, since they succeed at slowing down agreements (Figure~\ref{fig:costs_sr}), reputation loss is also slower. 

\begin{figure}
\centering
\includegraphics[width=0.66\linewidth]{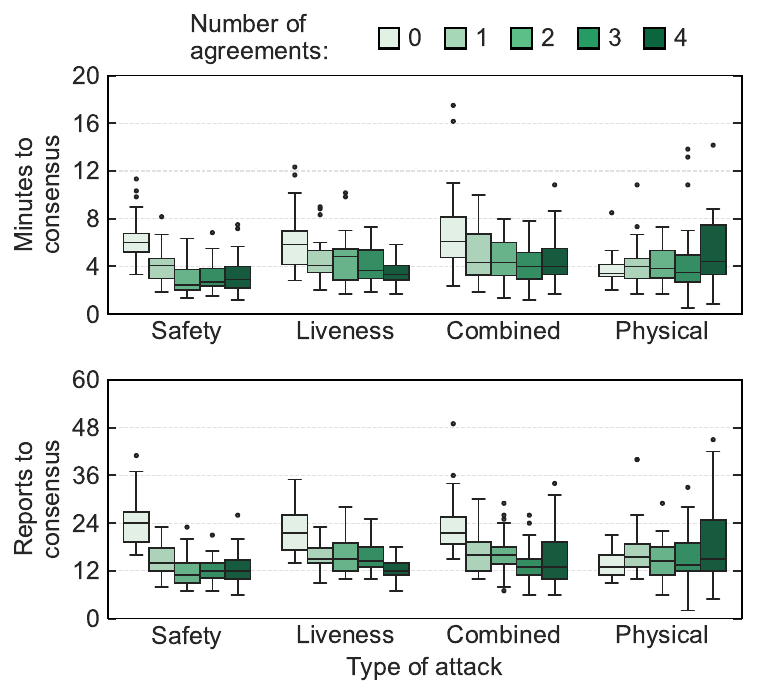}
\caption{Long-run autonomous recovery from faults, showcasing how costs reduce over time as agreements are reached.}
\label{fig:recovery_lr}
\end{figure}

Figure~\ref{fig:recovery_lr} shows that the cost of reaching agreements decreases over time, as the reputation system gradually mitigates the negative impact of the attackers. After approximately five agreements, most faulty robots have acquired low reputations. This results in a reduction of the average time to consensus, from 6 minutes for the first agreement to 3.7 minutes. The same trend is observed in the number of reports, which decreases from an average of 22 to just 13. These results highlight Swarm Oracle's self-healing ability, where it autonomously returns to an efficient operating state.

\section{Discussion}
\label{sec:discussion}

Our short-run experiments show that attacks by one or two robots are effectively neutralized with only a modest increase in consensus time (below $50\%$). However, when three or four attackers are present, the time to achieve consensus increases sharply. This is most evident when four Byzantine attackers are present since even a single unintentional fault in another robot can delay consensus or potentially lead to an incorrect agreement, if the faulty observation aligns with the attackers'. While incorrect agreements with $4$ attackers did not occur in the short-run experiments, it did happen in 5 different long-run experimental trials.
Nevertheless, once a first correct agreement is reached, the redistribution of reputation tokens reduces the impact of the attackers, making their success increasingly unlikely. This self-healing property was showcased by the long-run experiments, which demonstrate that the Swarm Oracle can recover over time and progressively nullify the effects of the attacks.

\paragraph{Scalability}
Achieving scalable fault-tolerance is one of Swarm Oracle's key advantages. Since the protocol's security relies on a global condition (in contrast with swarm robotics' local ones), larger swarms can tolerate a greater absolute number of faulty robots: While a swarm of 12 robots tolerates 4 faulty robots, a swarm of 48 tolerates 16. If we assume that 1 in 10 robots are faulty, the smaller swarm can withstand an additional 3 faulty robots, in comparison with 11 for the larger swarm. Larger swarm sizes also make attacks more challenging: attackers must coordinate a larger number of robots, which becomes increasingly difficult given the network size and spatial distribution of the robots.

Hardware costs also scale well, since the number of reports increases linearly with the number of robots, and communications between robots are local and very sparse. These costs are further reduced by our use of a cost-efficient blockchain protocol and dual-layer communications (infrared broadcasts for peer identification, and Wi-Fi for exchanging reports and block metadata). In Methods, we describe this technical setup in detail and show that costs in terms of onboard storage, computation, and communication are manageable. Previous research corroborates this claim with a comprehensive cost analysis with swarm sizes between 8 and 24 robots, and with up to 120 simulated robots~\cite{StrPacDor2023:sciencerobotics}.

The main scalability challenge arises from the requirement that a supermajority of the robots participate in consensus (more precisely, the holders of $\frac{2}{3}$ of the reputation tokens). This becomes costly when robots and observation areas are either widely dispersed, leading to large physical costs to generate reports, or overly concentrated---potentially leading to physical interference between robots. To mitigate this, Swarm Oracle could be deployed in smaller committees within a larger swarm. The committee size can be tuned to match application requirements, enabling fault-tolerance to scale in accordance with the levels demanded by the application. Our implementation also includes the possibility of parallelization of consensus (through the $K$ parameter), improving scalability by enabling robots to work on multiple consensus proposals at a time.

\subsection{Possible vulnerabilities and mitigation}

\paragraph{Denial-of-Service}
The deposit quota parameter $K$ limits the number of pending proposals, making it possible for attackers that continuously report incorrect observations to occupy all available slots. 
In theory, liveness is not compromised since asynchronous communication prevents attackers from reporting at the exact time a slot becomes available.
However, an attacker could willingly sacrifice all of its reputation to continuously attempt the attack, since reporting only requires a deposit proportional to the robot reputation. One solution is to not allow low-reputation robots to vote, effectively excluding them from Swarm Oracle until they are repaired. Alternatively, a minimum reputation could be required to open new proposals, allowing robots with low reputation to still participate in open proposals and regain reputation through useful contributions. 

\paragraph{Copycats}
To ensure the decentralized and trustless execution of the Swarm Oracle smart contract, it is required that observations, as well as robot reputations, are public so that they each compute the same weighted average. However, this transparency introduces a vulnerability: a malicious robot could wait to receive reports from others, and copy or average the observations from the most reputable peers to unfairly gain reputation. This type of attack is known in decentralized finance~\cite{barczentewicz2023defibots}, where trading bots copy the transactions of the most successful traders. 
Since hiding the reputations or observations is not possible, a potential solution is to instead conceal the votes using cryptography~\cite{VanPacStr-etal2023:SciRep}. In this approach, robots submit encrypted votes, which are only revealed once the proposal reaches its deposit threshold (after which no more reports are accepted). Robots that fail to reveal their vote would be penalized (e.g., by losing their deposit), and if a missing reveal is decisive for the outcome, the round is nullified.

\paragraph{Unobservability}
Byzantine robots may generate proposals based on fictitious observations, which are not present in the environment. In such cases, simply relying on robots to report their own observations is insufficient, as unobservable proposals may remain pending. In the worst case, if all available proposal slots are unobservable, Swarm Oracle could reach a deadlock. The simplest mitigation strategy is to automatically reject proposals that remain pending after a fixed number of minutes, blocks, or reports. Another is to have the robots vote negatively if they fail to observe anything matching the proposal within a reasonable time.
In our experiments, we adopted a variant of this approach: robots attempt to \textit{validate} existing proposals by searching the environment for a fixed amount of time, and then travel towards the observation that is the closest match. The resulting report includes the index of the proposal they aim to validate. If the report is not matched to the intended proposal by the clustering algorithm, then the reputation deposit is counted until the $\frac{2}{3}KT$ threshold is reached, after which the proposal is dropped.

\paragraph{Physical Attacks}
A key aspect of mobile robotics systems is that robot interactions occur not only through communication, but also physically. While our protocol protects against attacks exploiting communication, physical attacks also warrant attention.
In our robotics context, physical force may be used to interfere with other robots or manipulate the environment, inducing faults in otherwise non-faulty robots to degrade system performance and avoid accountability. Our experiments showed how physical attackers caused degradation comparable to that of combined attackers, but lost reputation at a slower rate.
Since physical interactions occur outside of digital communications, ensuring fault-tolerance remains an open problem. 
Interestingly, one solution could be to use Swarm Oracle itself to let the robots report observed physical attacks and agree on the identity of the attackers. These robots could then be permanently expelled from Swarm Oracle, and the information could be passed to law enforcement agencies or to the public blockchain, to hold robot stakeholders accountable.

\subsection{Swarm Oracle Deployment}

We deployed the Swarm Oracle on a group of 12 robots, each self-hosting an Ethereum~\cite{But2014:techreport} instance. Although a blockchain is not strictly necessary, Swarm Oracle requires methods to ensure an ordering of the reports, track reputation tokens, and prevent double-spending attacks (in which one robot could use its reputation tokens multiple times). Additionally, the robots need to synchronously and correctly execute the deterministic logic that underlies the Swarm Oracle protocol and its reputation system. These requirements are naturally fulfilled by blockchain smart contracts, which are widely available through open-source blockchains.

Blockchains also help to reduce the communication overhead and storage requirements by aggregating information into blocks generated at specified time intervals (every 10 seconds on average in our experiments).
Communication is very efficient since robots exchange reports through sparse and local gossip, and blocks require minimal bandwidth as they only contain essential metadata and the ordering of reports. Finally, the linear structure of the blockchain results in predictable storage growth, and the immutability of the cryptographic links between blocks allows robots to safely prune historical data to further save space. 

Swarm Oracle could also be deployed using other distributed ledgers, such as directed acyclic graphs~\cite{TraRamGow-etal2019:ledger} or blocklists~\cite{WarvonTro-etal2023:aamas}, or through protocols tailored for swarm robotics, such as SwarmMesh~\cite{swarmmesh} or Buzz~\cite{buzz}. However, these alternatives do not provide the security and convergence guarantees of blockchain-based protocols, and other potential advantages are yet to be demonstrated in practice. 

Another alternative is to deploy the Swarm Oracle smart contract on a public blockchain~\cite{DorPacReiStr2024naturereviewstech}. 
This reduces the communication and computational burden on the robots, while leveraging the decentralization and automation of smart contracts hosted on public blockchains. We tend to favor the local approach, where the Swarm Oracle protocols are maintained by the robots through a private blockchain, and only the consensus results are periodically uploaded to a public blockchain. This strategy preserves autonomy and fault-tolerance, while minimizing reliance on external connectivity and reducing transaction costs on public blockchains.

Swarm Oracle empowers robots from different stakeholders to pool their capabilities to provide reliable and transparent data as a service. Unlike traditional swarms, Swarm Oracle protocols are designed to withstand adversarial behaviors, enabling collaboration without the need for trust. The data gathered by Swarm Oracle can be provided to external applications, but it can also be employed internally by the swarm. Recent research increasingly explores how robot swarms can employ blockchains to support self-organization, governance and economic participation, while identifying the oracle problem as a fundamental challenge~\cite{DorPacReiStr2024naturereviewstech, Cas2018:ftc, PenKerSalFuYuWes2023}. Swarm Oracle can also be used to securely update distributed data structures maintained by the robots, other than blockchain, such as tuple-spaces~\cite{pinciroli2016tuple, denicola2020stigmergy, swarmmesh}, environment maps~\cite{kegeleirs2021swarmslam, lajoie2024swarmslam, MultiMap_IEEEproc}, coordinate systems~\cite{jones2024spatial, pluhacek2025coord}, directed acyclic graphs~\cite{TraRamGow-etal2019:ledger, KerPenWes2023:iotjournal}, blocklists~\cite{WarvonTro-etal2023:aamas}, and Turing-complete state machines~\cite{buzz}.

\section{Methods}

\label{sec:methods}
\subsection{The Swarm Oracle}
Our Swarm Oracle consists of a group of robots, denoted as $\mathcal{S}$. The robots, operating in a real-world environment, individually collect and process data using their own sensors and computational resources. The data obtained by the robots (e.g., readings from their sensors) forms an observation space, denoted as $\Omega$.
The goal is to develop a consensus protocol that lets robots agree on values in $\Omega$ while meeting the safety and liveness requirements: the protocol always produces correct and consistent agreements (safety), and every non-faulty robot eventually decides on a value (liveness). The {\it consensus set} $\mathcal{L} \in pow(\Omega)$ is the set of all agreements reached by the robots.

As an example, if the robots' observations consist of GPS coordinates marking the positions of valuable resources, then $\Omega = \mathbb{R}^2$, and $\text{pow}(\Omega)$ represents any possible set of positions on the Earth's surface. In this example, the goal of the consensus protocol is to enable the robots to continuously update the set $\mathcal{L}$ with accurate resource positions.

\subsubsection{Byzantine faults and attacks}

Robots operating in the real-world are susceptible to a wide range of potential faults. While some faults can be anticipated and mitigated through careful design of the robots, their behaviors, and their communication protocols, others cannot: a robot may develop faults causing unpredictable and erratic behavior, or multiple robots may collude to attack the system if controlled by a malicious adversary. Such faults are commonly referred to as Byzantine faults~\cite{lamport1982bft}.

We consider a Byzantine fault model because distinguishing unintentional from malicious faults is often impossible: a robot that fails to send observations due to broken wheels is indistinguishable from one that deliberately stops, and a robot that sends incorrect observations because of a dirty sensor is indistinguishable from one that does so to mislead the consensus. Malicious robots may, however, coordinate their actions to carry out stronger attacks, timing them to exploit faults present in other robots.
Therefore, the developed protocol must be \emph{Byzantine fault-tolerant} to guarantee correct and continuous operation in real-world deployments, even in the presence of malicious attacks and unintentional faults. 

Faults can also differ significantly in terms of duration: robots may experience temporary faults and subsequently recover, but they can also become permanently disabled or remain under an adversary's control for extended periods of time. Since the multi-robot system is assumed to operate autonomously, the protocol should include mechanisms to mitigate the long-term negative effects of permanent or persistent faults, which could otherwise degrade performance and ultimately lead to system failure.
To address this issue, we integrate a \emph{reputation system} into the protocol that weighs each robot's contributions based on its performance, eliminating the need for a repair technician or external authority during operation. Importantly, our reputation system does not exclude faulty robots: they are allowed to continue participating and may regain reputation if they recover from their faults.

\subsubsection{Protocol}

The number of robots in the Swarm Oracle is $N = |\mathcal{S}|$, and each robot has an associated digital identity $i = 1, \dots, N$, which controls $t_i$ reputation tokens. The reputation tokens of each robot, and consequently the total amount of reputation tokens $T = \sum_{i=1}^{N} t_i$, are globally known.

Let's initially assume that each robot can obtain perfect observations, for instance, in the case their sensors are noise-free. Even though this condition is not realistic and will be later relaxed, it helps to lay down and understand the functioning of our oracle consensus protocol.
Robots locally broadcast structured reports, which are disseminated across the swarm. A report $r$ contains an observation $r.o \in \Omega$, the robot identity $r.i$, a deposit of reputation tokens $r.d$, and a vote, signaling acceptance or rejection, $r.v$.
Through the blockchain, Swarm Oracle ensures a global ordering of the reports and that they are structurally valid, creating an array $\bm{r}$ of ordered reports. Then, through the Swarm Oracle smart contract, the following deterministic logic is applied to the array $\bm{r}$:

\begin{enumerate}
\item The reports are grouped according to their observations. Each group of observations generates a \textit{proposal}.
\item Once a proposal reaches a cumulative value of deposits greater than $\frac{2}{3}\, T$, the voting decision takes place.
\item An absolute majority is required to determine whether the proposal is accepted or rejected using the identities, deposits, and votes in the reports associated with the proposal. 
\item A reputation system redistributes the deposited tokens based on the outcome of the voting round.
\item The reports that have been used in the voting round are removed from $\bm{r}$.
\item If the proposal is accepted, it is added to the consensus set $\mathcal{L}$.
\end{enumerate}

\paragraph*{Byzantine fault-tolerance} 
It is important to understand the rationale behind requiring a $\frac{2}{3}$ quorum of the total reputation tokens in order to validate a proposal.
This value is optimal because it preserves both liveness and safety, as long as no more than $\frac{1}{3}$ of the reputation tokens are owned by faulty robots. Any change to this value would decrease the Byzantine fault-tolerance of the system:
By increasing it beyond $\frac{2}{3}$, the Byzantine robots could cause a deadlock with less than $\frac{1}{3}$ of the reputation tokens by simply not sending reports. 
Conversely, by decreasing it, the faulty robots could force incorrect agreements with less than $\frac{1}{3}$ (i.e. half of $\frac{2}{3}$) of the reputation tokens by obtaining the absolute majority stake in a voting round.

\paragraph*{Parallelization} 
In some applications, most notably those in swarm robotics, it may be inefficient to require a $\frac{2}{3}$ supermajority to reach agreements, since some robots may be slower than others when submitting their votes. To improve efficiency by enabling the robots to work in parallel, we introduce the \textit{deposit quota} parameter, denoted $K \in \mathopen]0,1\mathclose]$, which establishes the number of reputation tokens that a robot $i$ must deposit with its reports ($r.d = K\, t_i$), as well as the cumulative deposits required before a voting round occurs ($\frac{2}{3}\, K\, T$). Note that this parameter indirectly regulates the maximum number of pending proposals: there can only be $\lfloor K^{-1} \rfloor$ pending proposals at a given moment, otherwise, there would not be enough circulating reputation tokens to reach the cumulative deposits requirement on all of them, and the system could come to a deadlock.

With this in mind, steps 1 and 2 in the logic above are adjusted as follows:
\begin{enumerate}
\item The reports are grouped according to their observations, up to a maximum of $ {\lfloor K^{-1} \rfloor}$ groups. Each group of observations generates a \textit{proposal}.
\item Once a proposal reaches a cumulative value of deposits greater than $\frac{2}{3}\, K\, T$, it enters a voting stage.
\end{enumerate}

When observation tasks can be efficiently performed in parallel---for example, when observations result from unpredictable events, such as finding an object during random walks in large environments---then a low value of $K$ may be desirable so that the robots contribute towards multiple proposals simultaneously, exploiting parallelization and reducing physical interference. 
Conversely, if observations require dedicated effort (e.g., searching a specific area), or when the interference between robots is low, then higher values for $K$ may be warranted, allowing robots to concentrate on validating existing proposals and potentially accelerating the consensus process. The deposit quota $K$ is, therefore, a parameter that can be tuned as a function of the oracle's application context.

\paragraph*{Dealing with noise}
To have the robots reach agreements despite noisy observations, the individual observations of each robot stored in the shared array $\bm{r}$ are aggregated, e.g., by applying filtering and/or averaging functions. 
Our proposed method employs a clustering algorithm to group robot reports based on a similarity score (e.g., a distance function). In doing so, the reports from robots will be matched to the closest proposal. If a report is not matched to existing proposals, a new cluster is created as long as the number of clusters does not surpass $\lfloor K^{-1} \rfloor$ (otherwise, the observation is dropped and the reputation tokens are refunded). An aggregation function is then employed to transform the reports in a cluster $\mathcal{C}_j$, $j \in \{1,\ldots,\lfloor K^{-1} \rfloor\}$ into a proposal $p_j \in \Omega$. 

The clustering algorithm, the aggregation function and their parameters are chosen as a function of the oracle's application context and, in particular, of the observation space. In our experiments, we employed incremental k-means clustering~\cite{pham2004incremental}, with a simple rule for incrementing the number of clusters: a new cluster is generated when the distance between a new observation and all existing proposals exceeds a threshold $R$. The aggregation function generates proposals that correspond to the average value of the observations, weighted by the deposits associated with each report in a cluster. In section~\ref{sec:app} we explore the effects of varying the clustering threshold $R$.

\paragraph*{Autonomous recovery from faults}
In computer science, faults are often treated as temporary events, making dedicated detection and mitigation unnecessary, with prevention and tolerance preferred instead~\cite{anderson1983realtime}. In robotic systems, however, various faults may occur that robots cannot automatically recover from. While in controlled environments, such as warehouses, fault detection may suffice---since faulty robots can be serviced by technicians or removed from operation---in autonomous deployments it is crucial to integrate recovery mechanisms that protect against the accumulation of faults, which could eventually lead to system failure. However, since fault detection processes are not infallible and may yield false positives~\cite{lynch1996distributed}, simply blocking robots identified as faulty from participating in the oracle is not advisable. Our approach to this problem introduces oracle \textit{reputation tokens}, assigned based on each robot's contributions to the oracle and used as weights for its future contributions. A robot that has incurred many faults will have a lower reputation and, consequently, a smaller impact on the oracle's outcome. However, robots are always allowed to continue participating in the oracle and can therefore rebuild their reputation.

\subsubsection{Reputation system}

After each voting round, robots receive or lose reputation tokens based on their contributions. As mentioned above, the purpose of this is to enable the system to autonomously recover from faults: without such a measure, robots that are frequently or persistently faulty could continue to negatively affect the system performance and, as faults accumulate on different robots over time, this could lead to a system-wide failure.

The voting decision occurs when the $\frac{2}{3}\, K\, T$ quorum is reached.
Given the assumption that faulty robots possess fewer than $\frac{1}{3}$ of the existing tokens, and since robots must deposit exactly $K\, t_i$ tokens with each report, the absolute majority of tokens in each proposal originated from non-faulty robots. This ensures that the outcome of the voting round is correct. The robots that vote favorably to this outcome should receive reputation tokens, while the robots that vote against it should lose tokens. This reallocation of reputation tokens improves the \textit{safety} of the protocol for subsequent rounds. Additionally, it is important to penalize robots that abstain from participating in the voting round. This can be achieved either by removing some of their reputation tokens, or by diminishing their relative reputation by issuing extra reward tokens to the robots that participated in the voting round. 
Doing so makes the system more efficient and less likely to come to a deadlock (i.e., improves its \textit{liveness}), since the most active robots gain assets relative to robots that vote less often, or malicious robots that abstain from voting.

Different applications of the oracle may require different choices in the design of the reputation system or its parameters. A common downside of introducing reputation systems in peer-to-peer networks is the risk of introducing vulnerabilities that allow malicious agents to illegitimately acquire reputation tokens~\cite{sabater2005reputation}.

In our experiments, the number of tokens earned by a robot for sending report $r$ is given by the reputation gain function $G$: 

\begin{equation}
\label{eq:reward}
G(r, \mathcal{C}_j) =
\begin{cases}
    \frac{I_c}{|M(\mathcal{C}_j)|}+\sum\limits_{r' \in \mathcal{C}_j \setminus M(\mathcal{C}_j)} \frac{r'.d}{|M(\mathcal{C}_j)|} , & \text{if } r \in M(\mathcal{C}_j), \\
    -r.d, & \text{otherwise}
\end{cases}
\end{equation}

where $r$ is a report in the cluster $\mathcal{C}_j$, $M(\mathcal{C}_j)$ is the set of reports that form the (weighted) absolute majority of the cluster, $I_c$ is the \textit{issuance constant}, the parameter which establishes how many reputation tokens are generated each time a voting round finishes. 
More specifically, $M(\mathcal{C}_j)$ is defined as the following set:
\begin{equation*}
M(\mathcal{C}_j) = \left\{ r \in \mathcal{C}_j \mid r.v = v^* \right\}
\end{equation*}

where \( v^* \) is the unique weighted majority vote such that
\begin{equation*}
\sum_{\substack{r \in \mathcal{C}_j \\ r.v = v^*}} r.d > \frac{1}{2} \sum_{r \in \mathcal{C}_j} r.d.
\end{equation*}


Equation~\eqref{eq:reward} states that robots sending reports in $M(\mathcal{C}_j)$, in addition to receiving their deposited tokens $r.d$, will also receive an equal share of the deposits of robots sending reports not in $M(\mathcal{C}_j)$. The robots sending reports in $M(\mathcal{C}_j)$ also receive an equal share of the $I_c$. newly issued tokens. This means that the supply of reputation tokens changes after each voting round: after the $j^{th}$ voting round, the new quantity of circulating reputation tokens is $T_j = T_{j-1}+I_c$. 
Note that as more reputation tokens are added to the system, the $\frac{2}{3}$ threshold applies to the new quantity of circulating tokens, and the weights associated with existing deposits are adjusted to match the new token supply. In our experiments, we use $I_c=\frac{T_0}{\lfloor K^{-1} \rfloor}$, where $T_0$ is the total amount of reputation tokens at the start of each experiment.

\subsection{Execution Environment}

To implement and test our oracle system, we deploy a private blockchain hosted and maintained by the robots. In this setup, the reports sent by the robots are stored as ordered transactions in the blockchain's distributed ledger, and the oracle's consensus protocol is executed via a smart contract.
Previous research has shown that various blockchain consensus protocols can be utilized in robotic systems for this purpose. Pacheco et al.~\cite{PacStrDor2020:ants} demonstrated that inexpensive robots can execute blockchain software. In their work, the robots hosted a private Ethereum blockchain running a Proof-of-Authority consensus. Other private blockchains, such as Hyperledger~\cite{SalPenWes2023}, have also been deployed on robots, offering different trade-offs that emphasize different aspects of multi-robot system operation. 

As an alternative to employing private blockchains, Swarm Oracle's oracle consensus protocol could be executed as a smart contract on a public blockchain~\cite{DorPacReiStr2024naturereviewstech}, provided that the robots eventually connect to an external network, or on tailored protocols for robotic systems (e.g., Buzz~\cite{buzz}) which are particularly appealing to hardware-constrained systems such as nanorobotics or minimalist robotics~\cite{pluhacek2025coord}.

In this work, the Swarm Oracle was executed by twelve Pi-puck robots, each operating as a node in a private Ethereum network running a Proof-of-Authority consensus. Proof-of-Authority is a simple yet effective permissioned protocol that provides the desired security and convergence guarantees for our robotic system~\cite{ekparinya2020attack,wang2022exploring}.

\paragraph{Blockchain}
Each robot functions as an Ethereum node that independently maintains a copy of a blockchain. The reports sent by the robots are ordered and stored as transactions in blocks, and are used as inputs to update the state of a smart contract---which encapsulates the logic behind our oracle consensus protocol. The robots employ Proof-of-Authority~\cite{poa:online} consensus to resolve conflicts and agree on new blocks, ensuring that all robots maintain a synchronized state for their locally hosted smart contracts. In contrast with Proof-of-Work~\cite{Nak2008:techreport} consensus, Proof-of-Authority does not require spending computational resources and is therefore suitable for robotics applications. When the participants are perfectly time-synchronized, Proof-of-Authority is Byzantine-tolerant up to $\left(\frac{n}{2} - 1\right)$ Byzantine participants~\cite{poa:online}. However, if the participants are not synchronized, this value drops to the theoretical limit of $\frac{n}{3}$~\cite{ekparinya2020attack, wang2022exploring}.

\begin{figure}
\centering
\includegraphics[width=0.66\linewidth]{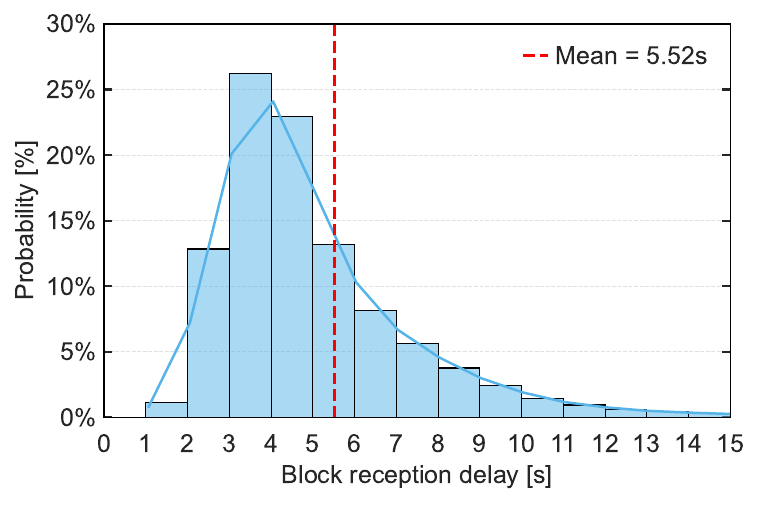}
\caption{Probability distribution of the block reception delay during short-run experiments (the time elapsed between the creation of a block and its reception by a robot). A total of 20\,002 unique blocks were produced, with 157\,561 block receptions logged by robots.}
\label{fig:blockdelay}
\end{figure}

The reception of transactions and blocks can be delayed due to sparse connectivity between the robots, which is influenced by the size of the environment and by the robots' limited communication range (approximately 15~cm in our experiments). Figure~\ref{fig:blockdelay} shows the probability distribution of block reception delays, showing that on average robots receive a block 5.52~seconds after it is produced, and that over 90\% of the blocks are received before 10~seconds. 
To allow time for transactions and blocks to synchronize across robots before producing another block, we set the block generation period to 10~seconds. This reduces forks (conflicting blockchain histories), minimizing bandwidth and processing overhead.
For a detailed analysis of block period impact on hardware costs, see~\cite{PacStrReiDor2022:ants}.


\begin{figure}
\centering
\includegraphics[width=0.66\linewidth]{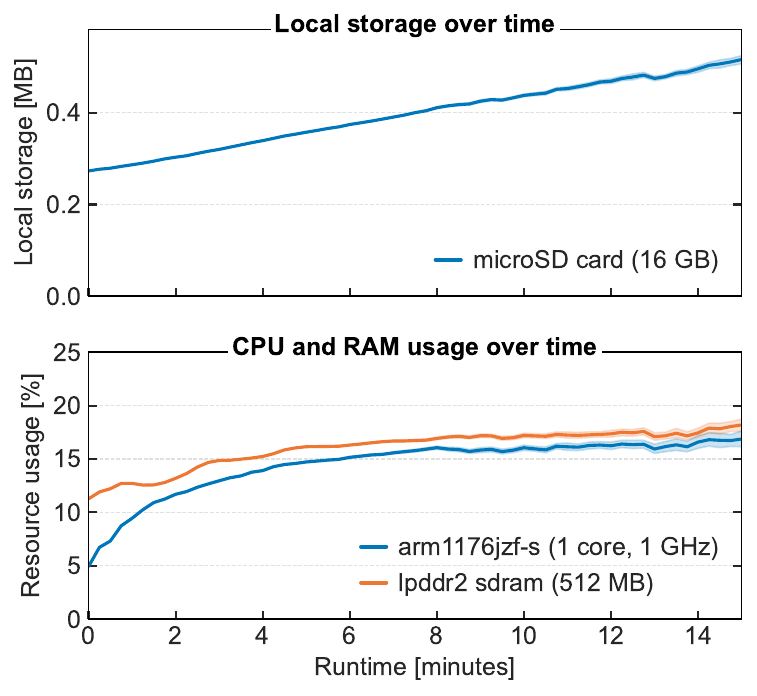}
\caption{Hardware resource usage by the Ethereum system process and data folder during short-run experiments. Storage, memory, and compute usage remain within the robots’ capabilities. The 95\% confidence intervals are tight, as the data was collected over a total experimental duration of 22~hours and 32~minutes of experiments, with robots logging values every 10~seconds.}
\label{fig:hardware}
\end{figure}

\paragraph{Robots}
We use Pi-puck robots~\cite{MonBonRae-etal2009:icarsc, MilJoyHil-etal2017:iros}, which have a Raspberry Pi Zero W onboard computer with a 16~GB SD card. This inexpensive hardware is sufficient to run most blockchain protocols that employ permissioned consensus protocols. Figure~\ref{fig:hardware} shows that the hardware demands of running Ethereum on the Pi-pucks, in terms of storage, memory and computing are very manageable.
The synchronization of the blockchain occurs over a Wi-Fi mesh network, where each robots communicates only with its physical neighbors. To find neighbors, robots use a range-and-bearing board to broadcast their IP addresses at a short range, and only when a robot receives this broadcast it allows a Wi-Fi connection to exchange blocks and transactions.

\begin{figure}
\centering
\includegraphics[width=0.5\linewidth]{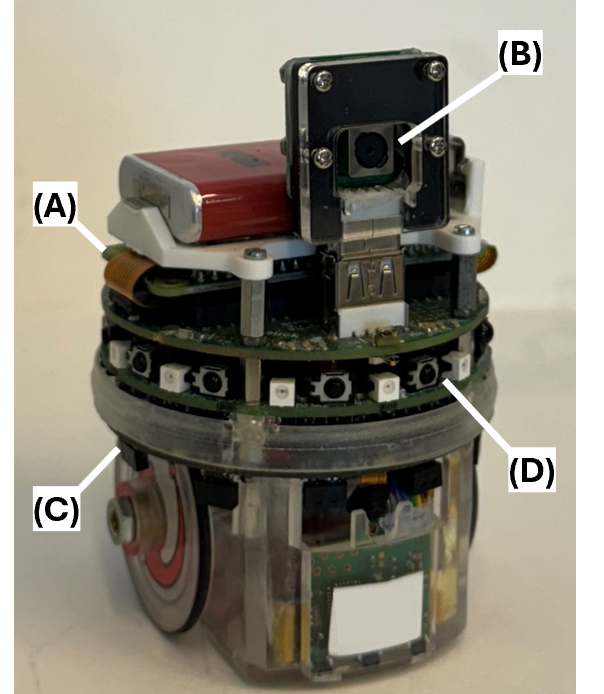}
\caption{The Pi-puck is a differential drive robot equipped with (A)~a Raspberry Pi Zero W with Wi-Fi capabilities, (B)~a front-facing camera, (C)~a ring of eight infrared sensors used for obstacle avoidance, and (D)~a range-and-bearing board that allows robots to communicate with local neighbors.}
\label{fig:pipuck}
\end{figure}

Robots perceive their environment using a ring of infrared sensors for obstacle avoidance, and a monocular camera for landmark detection and AprilTag recognition.
Figure~\ref{fig:pipuck} illustrates the Pi-puck robot and highlights its main sensors and actuators.

\paragraph{Simulator}
In addition to experiments with real robots, we provide extensive results obtained in simulation. The environment and the robots are reconstructed on a 1-to-1 scale in the physics-based simulator ARGoS~\cite{argos}. The controller of each simulated robot, which is identical to that of the real robots, communicates with Ethereum node instances that are executed inside Docker containers (one for each robot). The simulated robots obtain their camera readings from the dataset collected during the real robot experiments, each randomly drawing from the dataset of a corresponding real robot, and the camera view angles and camera occlusions are tuned to closely match those observed on the real robots. In this way, the simulated results provide a highly accurate extension of the results obtained with the real robots. 

\paragraph{Algorithm}
The Swarm Oracle's fault-tolerant protocol and reputation system are implemented through Algorithm~\ref{alg:scmc}, deployed as an Ethereum smart contract. The smart contract provides two functions for oracle participants: \texttt{Report}, which allows them to send reports by depositing the required amount of reputation tokens, and \texttt{Query}, which returns all accepted and pending proposals. 

The \texttt{Report} function is executed by sending blockchain transactions that encode the required deposit and report data. Once these transactions are added to the blockchain ledger, the state of the smart contract is updated. 
The \texttt{Query} function retrieves information from the current state of the smart contract, which is locally stored on each robot; this operation does not require a transaction.

The smart contract additionally employs four internal functions: a clustering function \texttt{Cluster($\bm{r}$)}, a distance function between two observations \texttt{Distance($o_1$,$o_2$)}, a function \texttt{Aggregate($\mathcal{C}_j$)} to aggregate the reports in cluster $\mathcal{C}_j$ into a proposal, and a reward function \texttt{Reward($r,\mathcal{C}_j)$} that implements the logic of the reputation system.
In this article, we employed incremental k-means clustering and an Euclidean distance function; the proposals correspond to the geometric center of the observations, weighted by the deposited reputation tokens; and the reward function is the one given in equation~\eqref{eq:reward}.
The modular design of these functions allows for easy replacement with other options better suited to different applications of the oracle. This design choice helps keep the protocol generic.

\begin{algorithm}[tbhp]
\footnotesize{
\caption{Oracle network consensus protocol}
\label{alg:scmc}

\SetKwInOut{Initialization}{Initialization}
\SetKwInOut{Parameters}{Parameters}
\SetKwInOut{Input}{Input}
\SetKwInOut{State}{State}
\SetKwInOut{Output}{Output}


\Input{$\mathcal{S}$, $\Omega$}
\Output{$\mathcal{L} \in pow(\Omega)$}
\Parameters{ $K$, $I_c$}
\State{$\bm{r}$, $T$, $\mathcal{L}$, $\mathcal{C}=\{\mathcal{C}_j\} $, $j \in \{1, \ldots, \lfloor K^{-1} \rfloor\}$} 
\Initialization{$\bm{r} \gets [ \, ]$, $T \gets T_0$, $\mathcal{L} \gets \emptyset$, $\mathcal{C} \gets \emptyset$, $j \gets 0$}

\SetKwComment{Comment}{$\triangleright$\ }{}
\SetKwFunction{Report}{Report}
\SetKwFunction{Query}{Query} 
\SetKwFunction{Balance}{Balance}
\SetKwFunction{Cluster}{Cluster}
\SetKwFunction{Distance}{Distance}
\SetKwFunction{Aggregate}{Aggregate}
\SetKwFunction{Reward}{Reward}
\BlankLine

\SetKwProg{amalg}{Procedure}{}{}
\footnotesize
\amalg{\Report{$r=(o,i,d,v)$}}{
    \If{$r.d<K\,\Balance(r.i) \vee r \in \boldsymbol{r}$}{
    \Comment{Reject due to insufficient deposit}
    \Comment{or due to duplicated reporting}
    Transfer $r.d$ reputation tokens to $r.i$ \; return 0 \;
    }
    $\boldsymbol{r} \gets \text{append}(\boldsymbol{r}, r)$ \;
    \Comment{Generate the set of clusters $\mathcal{C}$}
    $\mathcal{C} \gets \Cluster(\boldsymbol{r})$ \;
    \For{$\mathcal{C}_j \in \mathcal{C}$}{

    \If{$\sum_{r\in C_j}r.d \geq \frac{2}{3}KT$}{
        \If{weighted majority agreement}{  
        $M(\mathcal{C}_j) \gets r: r \in$ absolute majority \;
        \Comment{Reward reporting robots}
            \For{$r \in \mathcal{C}_j$}{
                $reward \gets G(r,\mathcal{C}_j)$ \;
                Transfer $reward + r.d$ reputation tokens to $r.i$ \;
            }
         \Comment{Update state variables}

            $T \gets T+I_c$ \;

            $\mathcal{L} \gets \mathcal{L}\bigcup \Aggregate(\mathcal{C}_j)$ \;
                         
            $\mathcal{C} \gets \mathcal{C}\backslash \{\mathcal{C}_j\}$ \;

            
        }
    }
    }
}

\SetKwProg{amalg}{Procedure}{}{}
\footnotesize
\amalg{\Query{
}}{
$proposals \gets \{\Aggregate(\mathcal{C}_j): \forall \mathcal{C}_j\in \mathcal{C} \}$ \;

$consensus \gets \mathcal{L}$ \;
    \Return{proposals, consensus} \;
}
}
\end{algorithm}

\subsection{Experimental Setup}
\label{sec:setup}
\subsubsection{Robot behavior}

\begin{figure}
\centering
\includegraphics{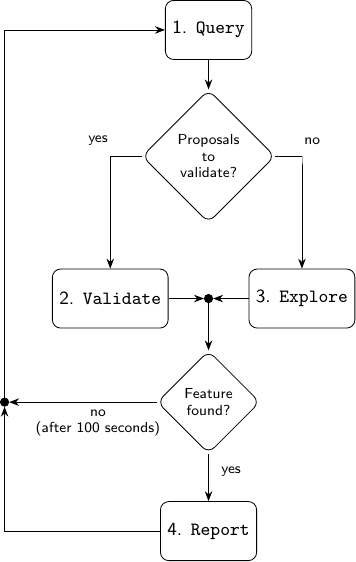}
\caption{The robots' state machine as a flowchart.} 
\label{fig:fsm}
\end{figure}

Our robots follow a routine that mainly consists of \emph{exploring} the environment to find landmarks and initiate new proposals, or \emph{validating} existing proposals in the smart contract. To achieve this, the robots detect the colors of landmarks in the environment and navigate towards them to measure their RGB values and read the AprilTags. The robots' state machine is illustrated in a flowchart in Figure~\ref{fig:fsm}.

\begin{enumerate}
    \item \texttt{Query} The robot queries the current proposals from the smart contract. If there are proposals it has not yet validated, it randomly selects one and transitions to the \texttt{Validate} state; otherwise, it transitions to the \texttt{Explore} state.
    \item \texttt{Validate} The robot searches the environment for landmarks and navigates to the one that is the closest match to the proposal to be validated. Upon arrival, it transitions to the \textit{Report} state, or returns to the \texttt{Query} state after $100$\,s.
    \item \texttt{Explore} The robot searches the environment for any landmark and then moves towards it. Upon arrival, it transitions to the \texttt{Report} state, or returns to the \texttt{Query} state after $100$\,s.
    \item \texttt{Report} Once the distance to the AprilTag attached to a landmark is smaller than a threshold, the robot reads the AprilTag and the average RGB value of the largest color contour in its camera vision. Then it sends a report through a blockchain transaction and returns to the \texttt{Query} state.
\end{enumerate}

\subsubsection{Parameters}
\paragraph{Clustering threshold}
The clustering threshold $R$ is an internal parameter of the k-means clustering algorithm that defines the minimum required distance between a new observation and existing cluster centroids (i.e., proposals) for a new cluster to be initiated.
In our experiments and real deployments, noisy camera readings can lead to faults, such as incorrect votes or robots failing to find proposals.
To examine the impact of faults caused by noisy measurements, we could keep a fixed $R$ and add noise or biases to the readings gathered by the robots. However, this approach would require artificial noise models, which are less interesting to analyze compared to the natural noise and color miscalibration already affecting the robots' readings. Instead we run a set of short-run experiments where we vary the value of $R$, to analyze the impact of faults caused by erroneous classifications due to noise. This is showcased in Figure~\ref{fig:short_R}. 

\paragraph{Number of Byzantine attackers} In the short-run experiments, we also vary the number of Byzantine attackers $f$. One of the main features of employing a global protocol is the certainty of correct agreements provided Byzantine robots remain below the $\frac{1}{3}$ threshold. This is shown in Figures~\ref{fig:error_sr} and~\ref{fig:costs_sr}, which shows that our method resists all attacks up to this point (4 Byzantine attackers, in a total of 12 robots).

\paragraph{Deposit quota}
The deposit quota $K$ establishes the percentage of reputation tokens that each robot must deposit when submitting a report. Indirectly, it regulates the maximum number of pending proposals: if $K=1$, only one proposal can remain pending, otherwise $\lfloor K^{-1} \rfloor$ proposals can remain open.
Notably, this has an effect on the dynamics of the token redistribution between the robots. In Figure~\ref{fig:reputation_lr} we can compare $K=\frac{1}{3}$ at the top, with $K=1$ at the bottom. With $K=\frac{1}{3}$, the redistribution of tokens is less steep (i.e., more time and agreements are required for Byzantine robots to lose tokens).

\paragraph{Issuance constant}
The issuance constant $I_c$ is employed in the first term of equation~\eqref{eq:reward} to generate new tokens that are rewarded to the robots that participated in the voting round. 
This generates an inflationary effect that penalizes robots that abstain from voting, and improves the liveness of the protocol. If the issuance constant is set to zero, the robots performing the attack on liveness are not penalized (Figure~\ref{fig:reputation_lr}). During a real deployment, this could lead to the accumulation of faults due to more robots becoming stuck or disabled, potentially compromising the protocol's liveness.

\subsubsection{Metrics and Baseline}

\paragraph{Consensus Error}
The consensus error is the Euclidean distance between the agreements generated by Swarm Oracle and the average value of all red observations made by non-attacking robots during the short-run experiments. Since the number of experiments was large, we consider the average values to be a good estimate of the RGB values of the landmarks' colors. A point cloud plot of these observations is shown in Figure~\ref{fig:rg_cloud}.

Using an estimated value is needed, since the true RGB values of the color panels are unknown (they depend on the camera sensors, their parameters and lighting conditions).
The lower the consensus error, the more precise our oracle was, i.e., the closer the robots' final agreement is to the average value.


\paragraph{Time to Consensus}
This is the time it took, in minutes, until an agreement on a ``valuable'' landmark is reached and recorded on a blockchain block. In the short-run experiments, it consists of the time to reach the first agreement, while in the long-run experiments we reset the clock each time an agreement occurs.

\paragraph{Reports to Consensus}
Similarly to time, this is the number of reports sent by the non-attacking robots before an agreement occurs. While time is an intuitive metric, the number of reports is invariant to factors such as the speed of the robots and how fast they can make observations, which are respectively related to, for example, the battery levels (which change as the experiments progress), and to the environment and observation spaces (which change depending on the scenario).

\paragraph{Baseline}
Distributed agreements in robotics are typically achieved through approximate local consensus, where each robot maintains an individual opinion that it shares with its neighbors. In such algorithms, the exact moment at which consensus is reached is not well defined, as agreement is determined by a threshold rather than by a discrete decision event. 
Additionally, robots are susceptible to attacks in which malicious robots exploit locality---for example, by flooding an isolated robot with incorrect reports---which makes it difficult or impossible to come up with safety guarantees. 
Due to these fundamental differences with respect to our exact consensus method, we do not consider approximate consensus approaches to be an adequate baseline. Instead, we opt for simplicity by comparing our approach to a baseline that consists of centrally collecting all observations made by the robots until the end of that experiment, and averaging them.





\bibliographystyle{unsrt} 
\bibliography{library_curated_manual}

\end{document}